\documentclass[runningheads]{llncs}
\usepackage[T1]{fontenc}
\usepackage{graphicx}
\usepackage{array}
\usepackage{multirow}
\usepackage{amsmath}
\usepackage{amssymb}
\usepackage{xcolor}

\usepackage{cite}
\begin{document}
\title{Vision Mamba for Classification of Breast Ultrasound Images}
%
%

\author{Anonymous}
\author{Ali Nasiri-Sarvi\inst{1} \and
Mahdi S. Hosseini\inst{1} \and
Hassan Rivaz\inst{2}}

\authorrunning{A. Nasiri-Sarvi et al.}

\institute{Department of Computer Science and Software Engineering (CSSE), Concordia University, Canada 
\and
Department of Electrical and Computer Engineering (ECE), Concordia University, Canada
}
\maketitle              

\begin{abstract}
Mamba-based models, VMamba and Vim, are a recent family of vision encoders that offer promising performance improvements in many computer vision tasks. This paper compares Mamba-based models with traditional Convolutional Neural Networks (CNNs) and  Vision Transformers (ViTs) using the breast ultrasound BUSI dataset and Breast Ultrasound B dataset. Our evaluation, which includes multiple runs of experiments and statistical significance analysis, demonstrates that some of the Mamba-based architectures often outperform CNN and ViT models with statistically significant results. For example, in the B dataset, the best Mamba-based models have a 1.98\% average AUC and a 5.0\% average Accuracy improvement compared to the best non-Mamba-based model in this study. These Mamba-based models effectively capture long-range dependencies while maintaining some inductive biases, making them suitable for applications with limited data. The code is available at {\color{magenta} \url{https://github.com/anasiri/BU-Mamba}}.

\end{abstract}

\section{Introduction}

Processing breast ultrasound images using deep learning can benefit from pretrained weights and a transfer learning paradigm, especially when dealing with smaller datasets \cite{yasaka2018deep}. Several factors influence the effectiveness of transfer learning. One crucial factor is the choice of the pretrained dataset \cite{zhuang2020comprehensive}, and another critical factor is the selection of the encoder architecture \cite{zhuang2020comprehensive}.

Convolutional Neural Networks (CNNs) \cite{CNN} and Vision Transformers (ViTs) \cite{vit,vit2} have been studied for breast ultrasound applications in works such as \cite{lazo2020comparison,al2019deep,gheflati2022vision}. Recently, a new family of encoders, Vim \cite{vim} and VMamba \cite{liu2024vmamba}, based on State Space Models (SSMs), has emerged in computer vision, leveraging the Mamba architecture \cite{mamba}. Although studies like UMamba \cite{umamba} have recently investigated the Mamba architecture for medical imaging dataset segmentation, its performance in breast ultrasound classification has yet to be explored.

In this work, we adopt these architectures from natural image processing, utilizing pretrained weights trained on ImageNet \cite{deng2009imagenet}, and compare them with common CNN and ViT architectures on two breast ultrasound datasets, BUSI \cite{busi} and B dataset\cite{BDataset}. We also conduct a statistical analysis to ensure that the performance improvements are significant. Our results show that Mamba-based architectures frequently outperform other encoder types with significant results. They are also never outperformed by the other encoders in terms of statistical significance.

\section{Related Work}

CNNs \cite{CNN} have shown tremendous capability in computer vision tasks, with AlexNet \cite{Alexnet} achieving breakthrough performance in the ImageNet Challenge \cite{deng2009imagenet}, leading to widespread adoption of deep learning models in the field with models such as ResNet \cite{resnet} and VGG \cite{vgg}. While CNNs excel at capturing local patterns, they are limited in capturing long-range dependencies due to their limited receptive field.

ViTs \cite{vit, vit2} have been proposed as an alternative to CNN models. They generally outperform these CNN models due to a lack of inductive bias present in CNNs, allowing ViT models to freely learn from the dataset using the attention mechanism \cite{attention,vaswani2017attention} and effectively capture long-range dependencies. However, this lack of inductive bias also means these models require more data and training to achieve similar performance to CNNs or outperform them.

Recently, a new family of architectures based on the Mamba model \cite{mamba} has emerged, leveraging State Space Models (SSMs) in deep learning \cite{s4}. These architectures, including VMamba \cite{liu2024vmamba} and Vision Mamba (Vim) \cite{vim}, show potential in various applications like video understanding \cite{video}, remote sensing \cite{remote}, pathology datasets \cite{Nasiri-Sarvi_2024_CVPR}, and point clouds \cite{pointcloud}. In medical imaging, Mamba-based models have demonstrated significant potential, particularly for segmentation tasks, as explored in \cite{umamba,liao2024lightm,ruan2024vm,wang2024mamba,xing2024segmamba}.

We aim to do a comparative analysis between the Mamba-based models compared to CNN and ViT architectures on two widely adopted breast ultrasound datasets BUSI \cite{busi} and  B dataset \cite{BDataset}. Since these datasets are small, our analysis uses transfer learning for the comparison. Similar work on comparative analysis for CNN and ViT has been explored for breast ultrasound datasets. For example, the impact of transfer learning on CNNs for breast ultrasound images is explored in \cite{lazo2020comparison,al2019deep,amiri2020fine}. In \cite{gheflati2022vision}, ViTs are compared to CNNs through transfer learning. Our goal is to extend these studies to Mamba-based models.  


\section{Preliminaries}
In this section, we provide an overview of the state space models. These preliminaries would help with understanding the Mamba model.

\subsection{State Space Models}

State space models are mathematical frameworks that describe continuous linear systems. The state equation defines how the state \( h(t) \) evolves as a function of the input \( x(t) \), as shown in Eq. \ref{eq:state_equation}.

\begin{equation}
    h_t = Ah_{t-1} + Bx_t
    \label{eq:state_equation}
\end{equation}

The output equation links the connection between the output, the hidden state, and the input based on Eq. \ref{eq:output_equation}.

\begin{equation}
    y_t = Ch_t+Dx_t
    \label{eq:output_equation}
\end{equation}



The parameters $A, B, \text{and } C$ are time-invariant and remain unchanged across the sequence in the S4 model \cite{s4}. This time-invariance allows using a global convolution to represent the sequential data, thereby avoiding the slowdown typically seen in recurrent modeling. The global convolution for a sequence of length $L$ is defined based on the Eq. \ref{eq:global_conv} ($\overline{CAB}$ is used to showcase discretized parameters): 


\begin{equation}
    \begin{aligned}
        y &= \bar{K} * x \\
        \bar{K} & \in \mathbb{R} := \mathcal{K}_L(A, B, C)  = (\overline{CB}, \overline{CAB}, ..., \overline{CA}^{L-1}\overline{B})
    \end{aligned}
    \label{eq:global_conv}
\end{equation}

Although this time-invariance accelerates processing speed, it restricts the model's ability to behave dynamically based on each input token, thus constraining its overall performance.

\subsection{Mamba}
In Mamba \cite{mamba}, selective state spaces are used where the state space parameters $B$, $C$, and $\Delta$ (a discretization parameter) are computed dynamically based on each input sequence, with $A$ as the only time-invariant parameter. This dependence on the input improves the model's ability to capture temporal variations and complex dependencies in the data. However, this dependence on the input prevents the option of using a global convolution, limiting the ability to utilize the parallel processing capabilities of GPUs due to the recurrent processing. 

Mamba uses a hardware-aware algorithm to increase the processing speed, considering that modern GPUs have two types of memory, SRAM and HBM, with the first being faster but having less capacity. Since the parameter $A$ is still time-invariant, they move it to the fast SRAM for the sequential processing. At each time step, they compute $B, C, \text{and } \Delta$ in HBM and move $B \text{ and } \Delta$ to SRAM to compute the state space recurrence. The state vectors are then moved to HBM to compute the output based on Eq. \ref{eq:output_equation} using the previously computed $C$. This hardware-aware algorithm makes Mamba's throughput a sequential model comparable to parallel transformers.

\section{Methods}
\label{sec:method}
 In this section, we explain two vision models based on Mamba Architecture. 

\subsection{Vim}
\begin{figure}
    \centering
    \includegraphics[width=\textwidth]{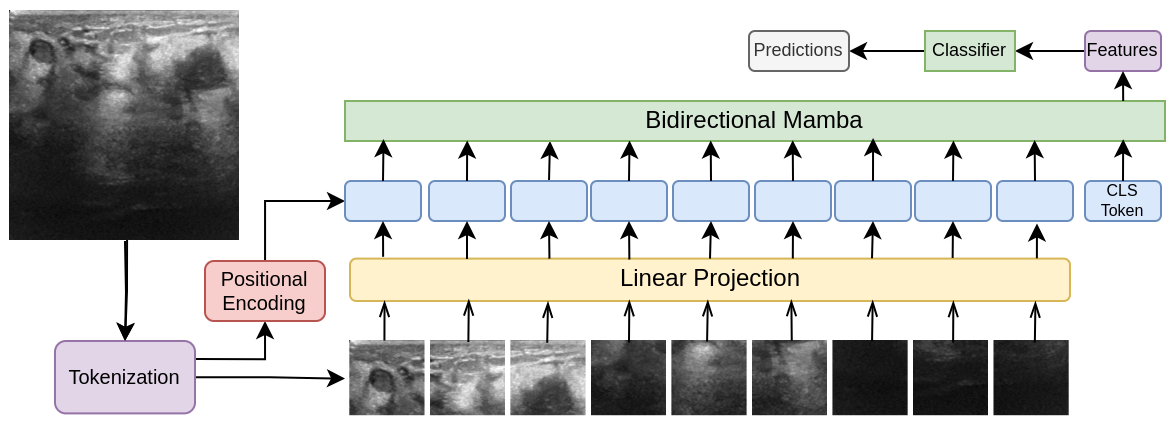}
    \caption{An over view of the Vim model}
    \label{fig:vim}
\end{figure}

In Vim, images are divided into smaller patches, each projected into a patch embedding. A bidirectional Mamba processes these patches by considering both previous and next tokens. Additionally, positional encoding is added to each token to retain spatial information about neighboring patches. The architecture is illustrated in Fig \ref{fig:vim}. Vim has similar processing to ViT models but uses Mamba-based blocks instead of attention-based Transformer blocks. 

\subsection{VMamba}
\begin{figure}
    \centering
    \includegraphics[width=\textwidth]{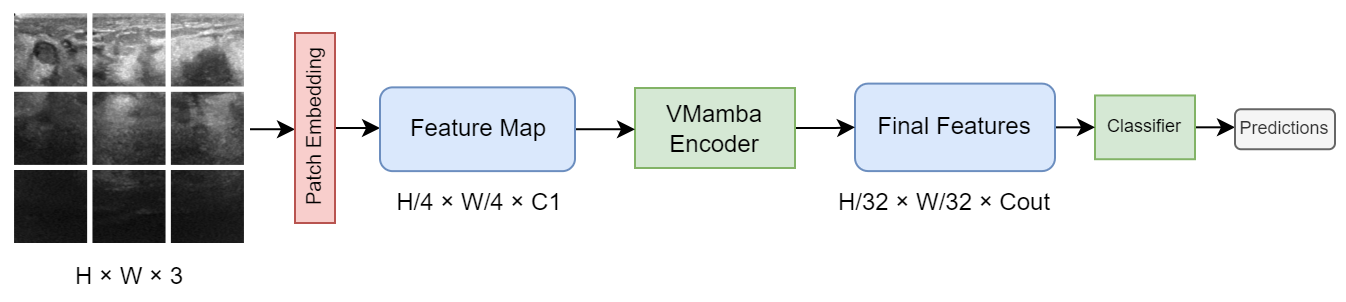}
    \caption{An over view of the VMamba model}
    \label{fig:vmamba}
\end{figure}

VMamba introduces ``2D Selective Scan'' to the Mamba's original scan. Furthermore, instead of breaking the image into tokens and processing each token similarly to ViT and Vim, the image patches are treated as feature maps and are processed using VSS blocks similar to CNN models, where the feature maps are down-sampled at each layer. The overall pipeline is demonstrated in Fig. \ref{fig:vmamba}.

\subsection{Model Comparison}
\begin{figure}
    \centering
    \includegraphics{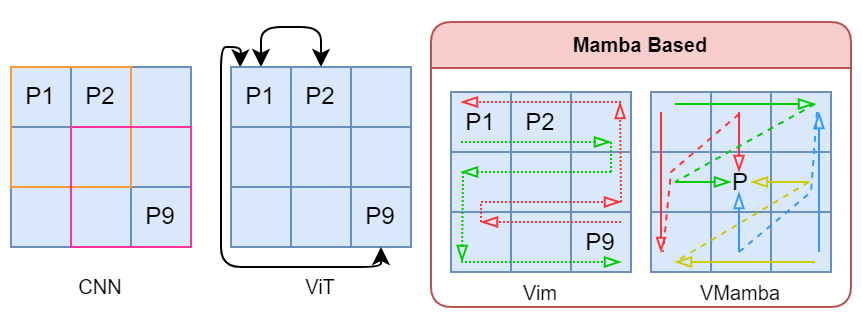}
    \caption{An abstract comparison between different architecture types.}
    \label{fig:arch_comp}
\end{figure}

Fig. \ref{fig:arch_comp} shows how different encoders process images. In CNNs, if patches P1 and P2 do not fall in the same receptive field, the model will struggle to capture long-range dependencies. In the ViT model, the attention mechanism would allow the model to process the relationship between P1 and P9. However, distinguishing between pairs like P1-P2 and P1-P9 relies primarily on positional encoding. As a result, the model needs more data and training to differentiate these pairs accurately and focus on close-range dependencies. This increased demand for training arises from the absence of the inductive bias, where neighboring patches are assumed to have similar content.

In Mamba-based models, the inductive bias is reintroduced through Mamba's sequential processing, similar to models like PixelRNN \cite{PixelRNN}. However, unlike PixelRNN, Mamba allows for long-range information processing and more efficiently utilizes GPUs. Consequently, Mamba-based models combine CNNs' inductive bias with ViT's long-range processing capability, offering the best of both worlds.

When comparing Vim and VMamba, Vim uses a bidirectional Mamba scan to achieve multi-directional feature extraction across the patches. In contrast, VMamba's 2D selective Scan employs the Mamba scan in four directions to capture more comprehensive and complex relationships between the image patches, enhancing the feature representation and in-context learning.


\section{Experiments}
This section summarizes the performance of various encoder types on different datasets. For our classification task, we use the BUSI \cite{busi} dataset (three classes of benign, malignant, and normal), the Dataset B of \cite{BDataset} (two classes of benign and malignant), and a combined BUSI+B dataset (three classes of normal, benign, and malignant). Each table reports the Area Under the Curve (AUC), Accuracy (ACC) metrics, and the number of parameters for each encoder. Since the experiment datasets are imbalanced, AUC is required as a more valuable metric. For BUSI and BUSI+B, which have three classes, the AUC is calculated individually for each class by comparing it against the other two classes. The final AUC for an experiment is then determined by averaging these individual AUC scores.

We use ResNet50 \cite{resnet} and VGG16 \cite{vgg} as the CNN-based models. For ViT models, we use ViT-ti16, ViT-s16, ViT-s32, ViT-b16, and ViT-b32 \cite{vit}, where the prefixes 'ti,' 's,' and 'b' denote tiny, small, and base model sizes, respectively, and the numbers indicate the patch size used (16x16 or 32x32 pixels). For Mamba-based models, we use Vim-s16 \cite{vim}, VMamba-ti, VMamba-s, and VMamba-b \cite{liu2024vmamba}. These Mamba-based models have the default parameters and definition as provided in the papers. We use pretrained weights of each of these encoders from ImageNet for transfer learning.  

We conduct each experiment across five training runs, using different seeds to minimize randomness in experiments. The data is split into training, validation, and test sets with respective splits of 0.7, 0.15, and 0.15. The validation set is used for early stopping; we select the checkpoint that shows the best performance on the validation set to prevent overfitting to the test set. Results are reported based on the test set's performance for each fold.


We assess statistical significance between two encoders using a paired t-test on their prediction. This analysis is conducted across all five folds, using the checkpoints from each model trained on the training set of a fold and subsequently tested on the test set.

\subsection{Results on the BUSI+B Dataset}

The classification performance of Mamba-based and ViT/CNN models is provided in Table \ref{tab:combined}. We ran our experiments five times and averaged the results to obtain mean and standard deviation measurements. Mamba-based models achieve higher or comparable results to ViT and CNN models. However, the high variance in the results and the small dataset size necessitate statistical significance analysis to compare the results. 

\begin{table}[h!]
\caption{Transfer Learning Results for BUSI+B datasets. The values of AUC and ACC are scaled between 0 to 100. The results are averaged over five runs.}
\centering
\begin{tabular}{|c|c|c|c|c|}
\hline
\multirow{2}{*}{\textbf{\begin{tabular} {@{}c@{}} Encoder\\Type \end{tabular}}} & \multirow{2}{*}{\textbf{Encoder}} & \multirow{2}{*}{\textbf{\begin{tabular} {@{}c@{}} \# Params\\   (Millions) \end{tabular}}} & \multicolumn{2}{c|}{\textbf{BUSI+B}} \\  \cline{4-5}

 &  &  & \textbf{AUC} & \textbf{ACC}  \\ \hline
\multirow{2}{*}{CNN} 
 & ResNet50 & 23.5 & $95.74 \pm 1.42$  & $87.66 \pm 2.04$   \\
 & VGG16 & 134.3 & $94.25 \pm 1.28$ & $85.82 \pm 1.49$  \\  \hline
\multirow{5}{*}{ViT} 
 & ti-16 & 5.5  & $94.19 \pm 1.74$ & $85.39 \pm 1.93$   \\ 
 & s-16 & 21.7  & $95.39 \pm 0.54$ &  $87.23 \pm 2.33$   \\ 
 & s-32 & 22.5 & $93.85 \pm 0.72$  &  $86.24 \pm 1.65$   \\ 
 & b-16 & 85.8 & $95.76 \pm 0.77$ & $88.51 \pm 2.67$   \\
 & b-32 & 87.5  & $95.51 \pm 1.53$  &  $86.52 \pm 3.23$   \\ \hline
Vim & s-16 & 25.4 & $95.84 \pm 0.96 $ & $87.38 \pm 3.22$    \\ \hline
\multirow{3}{*}{VSSM} 
 & VMamba-ti & 29.9  & $95.71 \pm 1.01$  & $89.36 \pm 2.33$   \\
 & VMamba-s & 49.4 & $96.12 \pm 0.75$ & $87.80 \pm 2.78$    \\
 & VMamba-b & 87.5 & $95.60 \pm 0.79$   & $88.51 \pm 2.22$    \\ \hline
\end{tabular}
\label{tab:combined}
\end{table}

\textbf{Statistical Significance Analysis}: To further evaluate the differences in performance between Mamba-based models and ViT/CNN models on the BUSI + B dataset, statistical tests are conducted using t-tests, and the threshold for the $p$-value is set to 0.05. The $p$-values were computed by comparing the accuracy of predictions between two models and testing for a zero mean difference.

This analysis reveals that VMamba-ti outperforms VGG16 (p-value: 0.004), ViT-ti16 (p-value: 0.003), ViT-s32 (p-value: 0.014), and ViT-b32 (p-value: 0.022), and the difference is significant. VMamba-b outperforms VGG16 (p-value: 0.037) and ViT-ti16 (p-value: 0.015). Additionally, ViT-b16 performs better than VGG16 (p-value: 0.044) and ViT-ti16 (p-value: 0.013). There are no other significant differences between pairs of models (the p-value is bigger than 0.05). This showcases that Mamba-based models perform similarly or outperform other encoders on the BUSI+B dataset with respect to statistical significance. 


It is important to note that while ResNet50, as shown in Table \ref{tab:combined}, achieves a higher mean AUC of $95.74$ compared to some of the other models, the statistical significance analysis indicates that this improvement is not significant. For example, when testing for ResNet50 against ViT-s32, the p-value is 0.27, even though the mean AUC and ACC for ResNet50 are better. Due to the experiments' small dataset size and inherent randomness, we report both mean and standard deviation for each metric, which makes direct comparisons between models more complex. Therefore, statistical analysis is necessary to reduce this randomness and better identify instances where the differences between pairs of models are statistically significant and not due to randomness in the experiments.

\subsection{Results on the BUSI Dataset}
The classification results for the BUSI dataset are provided in Table \ref{tab:BUSI}. We ran our experiments five times and averaged the results to obtain the mean and standard deviation measurements. The results suggest that the Mamba-based models outperform the other encoders on the B dataset.
\begin{table}[h!]
\caption{Transfer Learning Results for BUSI dataset. The values of AUC and ACC are scaled between 0 to 100. The results are averaged over five runs.}
\centering
\begin{tabular}{|c|c|c|c|c|}
\hline
\multirow{2}{*}{\textbf{\begin{tabular} {@{}c@{}} Encoder\\Type \end{tabular}}} & \multirow{2}{*}{\textbf{Encoder}} & \multirow{2}{*}{\textbf{\begin{tabular} {@{}c@{}} \# Params\\   (Millions) \end{tabular}}} & \multicolumn{2}{c|}{\textbf{BUSI}} \\  \cline{4-5}

 &  &  & \textbf{AUC} & \textbf{ACC}  \\ \hline
\multirow{2}{*}{CNN} 
 & ResNet50 & 23.5 &  $93.23 \pm 1.93$ & $85.64 \pm 2.72$  \\
 & VGG16 & 134.3   & $93.69 \pm 2.18$ & $85.47 \pm 4.59$   \\  \hline
\multirow{5}{*}{ViT} 
 & ti-16 & 5.5 & $93.52 \pm 3.07$ & $85.98 \pm 4.48$   \\ 
 & s-16 & 21.7  & $93.60 \pm 4.04$ & $86.50 \pm 4.44$ \\ 
 & s-32 & 22.5 & $93.59 \pm 2.59$ & $84.10 \pm 3.81$ \\ 
 & b-16 & 85.8 &  $94.11 \pm 2.12$ & $87.18 \pm 2.70$ \\
 & b-32 & 87.5  & $94.10 \pm 1.23$ & $85.98 \pm 1.39$   \\ \hline
Vim & s-16 & 25.4 &   $95.63 \pm 1.66$ & $87.86 \pm 2.72$   \\ \hline
\multirow{3}{*}{VSSM} 
 & VMamba-ti & 29.9   & $95.28 \pm 1.89$ & $88.55 \pm 1.67$  \\
 & VMamba-s & 49.4  & $94.48 \pm 3.48$ & $87.18 \pm 4.15$  \\
 & VMamba-b & 87.5 & $94.67 \pm 2.53$ &  $89.06 \pm 3.72$  \\ \hline
\end{tabular}
\label{tab:BUSI}
\end{table}

\textbf{Statistical Significance}: 
We performed $p$-value tests on the BUSI dataset. The analysis demonstrates that ViT-b16, VMamba-s, and Vim-s outperform ViT-s32 (with a p-value of 0.036, 0.027, and 0.011, respectively). VMamba-ti model outperforms ResNet50 (p-value: 0.032), VGG16 (p-value: 0.024), and ViT-s32 (p-value: 0.002). The VMamba-b model notably surpasses ResNet50 (p-value: 0.015), VGG16 (p-value: 0.006), ViT-ti16 (p-value: 0.018), ViT-s16 (p-value: 0.022), ViT-s32 (p-value: 0.0001), and ViT-b32 (p-value: 0.029). There are no other pairs with p-values smaller than 0.05. These results indicate that the Mamba-based models frequently performed at or above the level of other models with respect to statistical significance.

\subsection{Results on the B Dataset}
The classification results for the B dataset are provided in Table \ref{tab:B}. We conducted our experiments five times and calculated the average and standard deviation of the results. VMamba-ti is the best model, outperforming all other encoders on average AUC and having a large margin on average ACC. Other Mamba encoders also perform competitively and often surpass other models. 

\begin{table}[h!]
\centering
\caption{Transfer Learning Results for B dataset. The values of AUC and ACC are scaled between 0 to 100. The results are averaged over five runs.}
\begin{tabular}{|c|c|c|c|c|}
\hline
\multirow{2}{*}{\textbf{\begin{tabular} {@{}c@{}} Encoder\\Type \end{tabular}}} & \multirow{2}{*}{\textbf{Encoder}} & \multirow{2}{*}{\textbf{\begin{tabular} {@{}c@{}} \# Params\\   (Millions) \end{tabular}}} & \multicolumn{2}{c|}{\textbf{B}} \\  \cline{4-5}

 &  &  & \textbf{AUC} & \textbf{ACC}  \\ \hline
\multirow{2}{*}{CNN} 
 & ResNet50 & 23.5  & $90.05 \pm 4.19$ &  $78.33 \pm 5.53$ \\
 & VGG16 & 134.3  & $88.24 \pm 7.10$ & $80.00 \pm 6.67$ \\  \hline
\multirow{5}{*}{ViT} 
 & ti-16 & 5.5  &  $80.90 \pm 9.98$ & $77.50 \pm 7.26$  \\ 
 & s-16 & 21.7    & $90.68 \pm 7.89$ &  $ 82.50 \pm 6.67$ \\ 
 & s-32 & 22.5 & $87.55 \pm 8.89$  & $79.17 \pm 9.50$  \\ 
 & b-16 & 85.8  & $88.32 \pm 6.24$ & $76.67 \pm 7.26$ \\
 & b-32 & 87.5  & $83.14 \pm 12.82$ & $77.50 \pm 12.25$  \\ \hline
Vim & s-16 & 25.4  & $87.42 \pm 9.83$ & $84.17 \pm 8.50$  \\ \hline
\multirow{3}{*}{VSSM} 
 & VMamba-ti & 29.9  & $92.66 \pm 9.07$ & $87.50 \pm 12.08$  \\
 & VMamba-s & 49.4   &  $88.70 \pm 9.30$ & $83.33 \pm 10.54$ \\
 & VMamba-b & 87.5  & $92.19 \pm 5.43$ &  $81.67 \pm 6.24$ \\ \hline
\end{tabular}
\label{tab:B}
\end{table}

\textbf{Statistical Significance}: 
Our statistical significance analysis for the B dataset reveals that ViT-s16 outperforms ViT-b16 (p-value: 0.034). Vim-s performs better than both ViT-ti16 (p-value: 0.032) and ViT-b16 (p-value: 0.019). VMamba-ti model shows remarkable performance by outperforming ResNet50 (p-value: 0.011), VGG16 (p-value: 0.028), ViT-ti16 (p-value: 0.004), ViT-s32 (p-value: 0.012), ViT-b16 (p-value: 0.003), ViT-b32 (p-value: 0.007), and even VMamba-b (p-value: 0.034). There are no other pairs of models with p-values smaller than 0.05. These results underscore VMamba-ti as the top performer among the models evaluated. Furthermore, there are no instances with statistical significance where non-Mamba-based models outperformed Mamba-based models.

\section{Discussion}
The overall performance on the three datasets demonstrates the advantages of using Mamba-based models for breast cancer ultrasound datasets. Additionally, Mamba-based models' ability to capture long-range dependencies while retaining some inductive bias makes them a suitable alternative to CNN and ViT models, especially in scenarios with limited data and resources. Furthermore, VMamba showed better performance overall compared to Vim. This could be due to 2D selective scan of VMamba, which allows better representation learning. 

One limitation of this study is that the statistical significance analysis is not considering the imbalance in the dataset due the multi-class analysis and complexity of comparing AUC for pairs of models. Despite this, some  Mamba-based models frequently have higher average AUC as provided in the experiments section. A more comprehensive statistical analysis, including the multi-class AUC, would provide stronger support for our results.

\section{Conclusion}

In this work, we conducted a comprehensive comparison of three families of vision encoders—ViT, CNN, and Mamba-based models—using the BUSI and B datasets and a combined BUSI+B dataset. Our evaluation included multiple runs, averages, and standard deviations, and we performed a statistical significance analysis for each experiment. Overall, Mamba-based models frequently demonstrated competitive performance compared to the other models, with statistically significant results. Additionally, they were never outperformed by any other encoder type regarding statistical significance.

\begin{credits}
\subsubsection{\ackname}
This project was funded by the Natural Sciences and Engineering Research Council of Canada (NSERC) and Fonds de recherche du Québec – Nature et technologies (FRQNT). We also extend our gratitude to Behnaz Gheflati for their valuable time.

\subsubsection{\discintname}
 The authors have no competing interests to declare that are relevant to the content of this article.
\end{credits}
%
%
%
%

\bibliographystyle{ieeetr}
\bibliography{references}





\end{document}